\documentclass[twocolumn,10pt]{article}
\usepackage[letterpaper,top=0.75in,bottom=0.85in,left=0.62in,right=0.62in]{geometry}
\usepackage[T1]{fontenc}
\usepackage{newtxtext}
\usepackage{amsmath,amssymb}
\usepackage[hyphens]{url}
\usepackage{graphicx}
\usepackage[round]{natbib}
\bibpunct{(}{)}{;}{a}{}{,}
\let\cite\citep
\usepackage{caption}
\usepackage{booktabs}
\usepackage{microtype}
\newcommand{\srepro}{\sigma_{\mathrm{repro}}}

\title{Decomposing Yes--No Bias in LLM Moral Judgment:\\ Stance, Position, Word, and Verdict}
\author{Haonan Huang\\[2pt] Princeton University, Princeton, NJ, USA}
\date{}

\usepackage{pdfpages}
\usepackage[hidelinks,pdfauthor={Haonan Huang},pdftitle={Decomposing Yes-No Bias in LLM Moral Judgment: Stance, Position, Word, and Verdict}]{hyperref}

\begin{document}

\maketitle

\begin{abstract}
Large language models (LLMs) increasingly issue binary verdicts---safety gates, judge decisions, value
probes---and the answer is read as the model's judgment. A settled judgment should be coherent across paired
phrasings of the same question: yes to ``do you approve?'' should mirror no to ``do you oppose?''. LLMs often violate this
coherence, and in a yes/no question the verdict, the answer word, and the printed position coincide, so a
conventional flipped-question audit yields one contrast that cannot say what moved. We introduce a crossed
psychometric battery that extends the flip to full $2{\times}2$ and $2{\times}2{\times}2$ crossings of question
polarity, printed order, and answer label, separating stance from position, word, label, and verdict-role
effects; an independent graded instrument estimates each model's stance $\theta$ with no binary format; and a
two-parameter fit (framing susceptibility $m$, decisiveness $s$) separates genuine robustness from saturation
that merely hides susceptibility. Applying the battery to twenty moral dilemmas---1.35 million trials over ten
frontier models from the Claude, GPT, and Gemini families in 23 reasoning configurations---we find response-form
bias widespread but never uniform. The largest apparent yes--no biases are almost purely surface (word and
position), while several of the newest models show the reverse: near-zero flip contrasts concealing verdict
attachment up to $+0.42$, a component no surface counterbalance can expose, replicated across label alphabets.
Requested reasoning reshapes these channels rather than removing them. Finally, when subjects state decisions in
free text and a separate model transcribes them, two judge families agree on 99.3\% of jointly scored transcriptions with zero
opposing committed labels, and the transcribed stance converges with the graded ruler (median $r=0.87$):
response form is part of the instrument, and the judgment can be measured apart from it.
\end{abstract}

\section{Introduction}

Language models are moving into roles where their answers carry consequences: they screen content, advise on
consequential questions, grade other models' outputs, and act as agents whose choices touch people who never
see a prompt. Whether these systems are aligned has therefore become a question answered by measurement, not
assumption~\cite{wallach2025measurement}---and the workhorse measurement is binary. Safety evaluations ask
whether an action is acceptable; moral batteries record approve or oppose; judge models return pass or fail. In
each case the elicited bit is read as the model's judgment.

That reading assumes a lot. Survey science spent decades learning that a recorded answer is a joint product of
respondent and questionnaire~\cite{schuman1981,tourangeau2000,krosnick1987}, and LLMs give the
lesson new force: answers move under changes of wording, order, and label that carry no logical
content~\cite{sclar2024quantifying,pezeshkpour2024large,zheng2024large,tjuatja2024exhibit}. The sharpest
instance in the value domain is the \emph{yes--no bias}~\cite{cheung2025amplified}: ask a model whether it
approves of a course of action, and separately whether it opposes it, and the answers too often agree---yes to
both, or no to both---far beyond human framing sensitivity, and stably across repeated generations. A model that answers this way has no readable verdict in any single question: an audit that asks once---as
most do---can pass or fail a system for reasons unrelated to its values.

Documenting the bias is not understanding it: the phenomenon is underidentified in the format that
reveals it. In a yes/no question three logically distinct things coincide in one token: the \emph{semantic
verdict} (endorsing or rejecting the proposition), the \emph{answer word} (``yes'' or ``no'' itself), and the
\emph{printed position} (which option appears first). When the flipped question moves the answer, any of the
three may have moved it---different failures with different fixes---yet the flip-and-compare audit yields one
contrast, unattributable among three causes; and since incidental phrasing also moves answers, a single pair of
forms estimates even that contrast noisily. Identification takes both: \emph{crossing}---balanced counterfactuals
flipping each factor independently---and \emph{families} of logically equivalent forms that separate the
stable bias from form-to-form noise.

This paper builds that instrument and applies it at scale: a graded ruler estimating each model's per-dilemma
stance $\theta$ with no binary format; a polarity $\times$ printed-order $2{\times}2$ that splits the
conventional contrast exactly into position and word terms; a polarity $\times$ label $\times$ order
$2{\times}2{\times}2$ that separately identifies stance, label, position, and \emph{verdict attachment}---a
preference for the affirmative verdict itself, which no surface counterbalance can expose; and a two-parameter
fit ($m$, $s$) that tells genuine robustness from saturation. Across twenty moral dilemmas, ten frontier models,
23 configurations, and 1.35 million trials, the diagnosis is that response-form bias is widespread but never the
same phenomenon twice: the largest apparent biases are almost purely surface, while several of the newest models
hide large verdict attachment beneath near-zero flip contrasts, and requested reasoning reshapes channels rather
than removing them. The remedy comes from the same battery: three readouts that avoid or counterbalance the binary code---the ruler, the
symmetrized stance, transcribed free text---converge on this panel: an evaluator can measure the channel and still extract the judgment.

\section{Related Work}

\paragraph{Moral and value evaluation.}
LLMs are evaluated as moral classifiers, respondents, and simulated populations across benchmark,
psychological, and social-scientific
settings~\cite{hendrycks2021aligning,jiang2021delphi,scherrer2023evaluating,nie2023moca,abdulhai2024moral,miotto2022who,santurkar2023whose,oh2025robustness,perez2023discovering},
making elicitation reliability itself an alignment question: an apparent value difference is uninterpretable
if an equivalent format would reverse it. Our goal is not moral adjudication or human
comparison but convergent and discriminant validity within a model under controlled
transformations~\cite{campbell1959,cronbach1955,jacobs2021measurement,bean2025measuring}.

\paragraph{Prompt sensitivity and response effects.}
LLM answers move with wording, option order, label choice, and survey
form~\cite{sclar2024quantifying,pezeshkpour2024large,zheng2024large,wang2023primacy,dominguezolmedo2024questioning,tjuatja2024exhibit,rottger2024political,zhu2023promptrobust,mizrahi2024state,rupprecht2025prompt,zhang2025yes,song2025negative},
echoing the human questionnaire-design canon~\cite{schuman1981,sudman1996,tourangeau2000,schwarz1999}. Most
robustness studies measure variation over prompts and report it as a consistency score, or average it away; a
calibration line instead estimates a content-free answer-token prior and corrects for it~\cite{zhao2021calibrate},
or rescores competing surface forms of the same answer~\cite{holtzman2021surface}.
Closest in method, full-factorial experiments crossing order, labels, and framing show that prompt architecture
itself induces artifacts and recommend aggregating over architectures~\cite{brucks2025prompt}; in this
conference series, a marginal preference for yes over no has been studied as an exploitable implicit
bias~\cite{du2025yesno} and corrected at the token level~\cite{bhatt2025yesno}. The
present contribution is structural: crossing the factors that make an exact split of the conventional contrast
possible, adding the neutral-label crossing that isolates verdict attachment, and measuring stance independently
of the disputed format. We use neutral component names---position, label, answer-word,
verdict-attached---because the factors are experimentally defined; a pull toward a yes/no token does not imply a
content-level tendency to agree~\cite{braun2025acquiescence}, and a verdict-attached component implies neither
failed logic nor sycophancy toward a user view, since none is expressed~\cite{sharma2024towards}.

\paragraph{Model-based evaluation.}
LLM judges convert open-ended generations into labels and carry position~\cite{zheng2023judging,wang2024large} and
verbosity~\cite{zheng2023judging} biases of their own. We therefore treat the judge as an instrument too---perturbed, repeated, and cross-checked across two
families with abstention preserved---and its task is semantic transcription of a stated decision, not fresh
moral adjudication.

\section{The Crossed Battery}

The design separates three things that a single benchmark score fuses: a \emph{battery} of crossed forms
posing the same decisions (the instrument, portable to other binary evaluations), a fixed
\emph{dilemma set} it runs on---each item designating one of its two courses as the positive pole,
here the cost--benefit (CBR) course defined below---and a \emph{panel} of models and inference configurations
it evaluates.

\subsection{The Battery}

\paragraph{Graded ruler.}
Each item's designated course and its complement are rated under 24 crossed unipolar conditions---numeric scale
($0$--$10$, $0$--$100$), anchor direction, wording, and semantic pole---each a separate stateless call. Ratings
normalize to acceptability $a\in[0,1]$, and
\begin{equation}
\theta_{ir}=\overline{a}_{\mathrm{action},ir}-\overline{a}_{\mathrm{complement},ir}\in[-1,1]
\label{eq:theta}
\end{equation}
is item $i$'s stance in configuration $r$, positive toward CBR. Cross-form incoherence $\srepro$ is the spread
of per-form stance estimates across logically equivalent forms after subtracting within-form sampling
variance, a gauge-R\&R-style decomposition~\cite{burdick2003}; generalizability $G$ is the between-item
variance share~\cite{cronbach1963}. Separately queried, stateless calls cannot coordinate, so agreement across scales, anchors, wordings, and
poles must come from a stable underlying disposition---convergent evidence in the
construct-validation sense~\cite{campbell1959,cronbach1955}---and what makes $\theta$ usable as a reference
coordinate.

\paragraph{Crossed binary forms.}
Every binary analysis runs on one primitive, the \emph{flip pair}: two forms that pose the same decision and
differ in one response factor. With $p_{+}$ and $p_{-}$ the pair's CBR-endorsement probabilities---cells grouped
so that in the $+$ member the factor's positive pole (yes token, first-printed slot, A label, affirmative role)
carries the CBR course---each dilemma yields
\begin{equation}
z=p_{+}+p_{-}-1,\qquad b=p_{+}-p_{-},
\label{eq:zb}
\end{equation}
both on the signed axis: the symmetric part $z$ estimates the stance (the response factor cancels) and the
antisymmetric part $b$ is that factor's pull (the stance cancels). The pairing convention is load-bearing (grouping approve-yes with oppose-yes instead would fold stance into
``bias''), and every channel below is a $b$ under it, the crossing's remaining factors balanced.

The matched crossing varies question polarity (approve/oppose, intended complements) and printed order (yes-first/no-first) over 12
matched form pairs per dilemma, each pair differing in exactly the verb token. At fixed order $o$ the polarity pair \emph{is} the answer-word channel's flip
pair---approve makes \emph{yes} the CBR answer, oppose makes it \emph{no}---so its bias $d_o$ is zero for a
respondent who treats the polarities as intended complements (one at genuine indifference answers each form at
chance, still contributing zero), and crossing print order splits the conventional contrast exactly:
\begin{equation}
d_{\mathrm{yf}}=
\underbrace{\frac{1}{2}(d_{\mathrm{yf}}-d_{\mathrm{nf}})}_{b_{\mathrm{position}}}
+\underbrace{\frac{1}{2}(d_{\mathrm{yf}}+d_{\mathrm{nf}})}_{b_{\mathrm{word}}}.
\label{eq:decomp}
\end{equation}
The position term flips with print order; the word term survives order-balancing. What this factorial cannot
resolve is whether its word term follows the printed token or the logical verdict, since yes/no carries both. The
neutral-label crossing breaks the tie: with A/B labels and independent variation of polarity, label mapping,
and semantic print order, its projections separately estimate stance, label pull, position pull, and
\emph{verdict attachment}---the residual preference for the affirmative versus negative verdict once every
surface factor is counterbalanced (the design's remaining interaction projections are reported in the
supplement\footnote{The supplement referred to throughout is appended after the references (pp.~10--65).}). Beyond A/B, the battery carries thirteen alternative label pairs---numerals, symbols, marks, and yes/no words
in four languages---with full polarity-crossed cubes for four of them (supplement). Estimates use complete
factorial cells with item-clustered bootstrap intervals. Throughout, the inferential unit is the dilemma:
intervals resample items---at most eighteen clusters per contrast---with forms, replicates, and configurations
fixed, so 1.35 million calls buy stable cell estimates, not degrees of freedom, and every interval is a nominal,
unadjusted 95\% interval for a prespecified displayed contrast. Eligibility, coverage, and sensitivities are in
the supplement.

\paragraph{Susceptibility and decisiveness.}
Because the ruler establishes a continuous within-model stance, the parsimonious model of a forced binary is the
canonical logistic link: a channel's flip pair responds as
\begin{equation}
p_{\pm}=\sigma\!\left((\theta\pm m)/s\right),
\label{eq:ms}
\end{equation}
so its $z$ and $b$ (Eq.~\ref{eq:zb}) follow from two parameters: $m$, the frame's horizontal offset
(\emph{framing susceptibility}, expressed in units of the $\theta$ coordinate), and $s$, the width of the transition (\emph{moral decisiveness}, a scale of its own: with small $s$ the response snaps from 0 to 1 at the model's
tie; with large $s$ it hedges). Averages are relative to the fixed dilemma set; $m$ and $s$ are identified chiefly by the dilemmas nearest
the tie. Fits are per configuration and channel by nonlinear least squares, with
item-bootstrap intervals, read within their saturation and degeneracy flags (supplement).

\paragraph{Free-response transcription.}
The last protocol removes the response code from the subject entirely: shown the scenario and the two courses,
the subject states a decision in free text---no scale, no yes/no, no A/B, no printed response options. A separate judge
model transcribes the committed outcome as CBR-oriented, Rule-oriented, or \emph{no clear commitment} (the
registered abstention category), with policy refusal recorded separately. The
transcription layer is itself audited---judge order swapped, judge labels neutralized and counterbalanced,
judgments repeated, the sweep replicated by a second judge family, and a blinded, hard-case author-coded audit
supplying a human check on the commitment boundary (supplement).

\begin{figure}[t]
\centering
\IfFileExists{figures/fig1_identification/fig1_identification.pdf}{%
\includegraphics[width=\columnwidth]{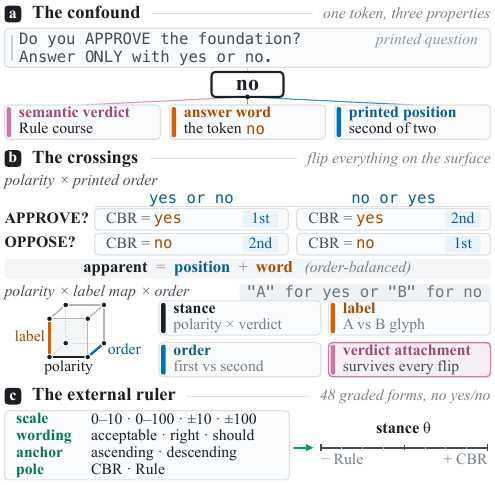}}{%
\fbox{\parbox{0.9\columnwidth}{\centering\vspace{6em}Fig.~1 --- in production\vspace{6em}}}}
\caption{\textbf{One yes/no answer is three measurements at once; crossing separates them.} Top: in a yes/no
form the semantic verdict, the answer word, and the printed position coincide in one token. Middle: polarity
$\times$ order splits the conventional contrast into position and an order-balanced word term; neutral A/B
labels crossed with polarity and order separately identify stance, label, position, and verdict attachment---the
component no surface counterbalance can remove. Bottom: the graded instrument crosses scale, anchor, wording,
and pole (48 collected forms; the 24 unipolar conditions define $\theta$, Eq.~\ref{eq:theta}) with no binary
format.}
\label{fig:identification}
\end{figure}

\subsection{Dilemmas and Stance Axis}

The battery covers 20 established moral dilemmas spanning sacrificial action, omission,
redistribution, and assisted dying; nineteen preserve their source vignettes verbatim, one is an explicitly
marked adaptation~\cite{koerner2023dilemmas,cheung2025amplified,maier2025code}. Each is oriented on one descriptive axis: one
course follows na\"{\i}ve cost--benefit reasoning (CBR), ``committing sacrificial harm or breaking a moral rule
for the greater good''; the other is the Rule course, consistent with following a moral
rule~\cite{cheung2025amplified}---roughly the utilitarian versus deontological poles, a gloss we use only here. Pole
assignments are frozen item metadata, decoupled from action versus omission; the story text never changes, only
the elicitation form. Sign conventions, stated once: stance-like
quantities ($\theta$ and the binary stance projections) are signed positive toward the CBR course;
response-channel components are signed on their own poles---toward the \emph{yes} token, the
\emph{first-printed} option, the \emph{A} label, the \emph{affirmative} role. Both classes are reported on the signed $[-1,+1]$ axis (choice
probabilities enter as $2P-1$), so a channel magnitude reads against a maximum pull of one---a component of
$0.2$ is a fifth of the largest possible pull; diagnostics and fitted parameters keep their own stated scales.

\subsection{Models and Contract}

The panel covers ten frontier models in 23 configurations (12 Claude, 9 GPT, 2 Gemini) spanning no, low, high,
manual, and adaptive reasoning; every one of 1,353,104 planned calls completed. Each trial is a fresh,
stateless, tool-free, single-turn request at temperature 1 with no system prompt---the two Gemini models alone
receive a disclosed one-line instruction enforcing a parseable format. Model version, request contract, and
serving date are part of the measured system. The primary panel is the same 18 dilemmas everywhere; two
veterinary items trigger provider-specific refusals and enter separately (details in the supplement).

\section{Results}

\subsection{The Ruler: A Coherent, Discriminating Stance}

\begin{figure}[t]
\centering
\IfFileExists{figures/fig_theta/fig_theta.pdf}{%
\includegraphics[width=\columnwidth]{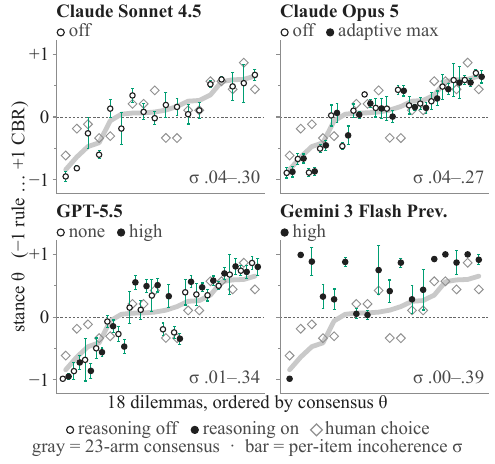}}{%
\fbox{\parbox{0.9\columnwidth}{\centering\vspace{6em}Fig.~2 ($\theta$) --- in production\vspace{6em}}}}
\caption{\textbf{Logically equivalent graded forms recover a stable within-model stance.} Per-dilemma $\theta$
on the common 18 dilemmas (whiskers: cross-form incoherence, not sampling CIs), ordered by the 23-configuration
panel mean (the gray ``consensus''; descriptive); open/filled markers contrast thinking off/on; diamonds are human choice rates
from the source battery ($2P{-}1$; context, not a target). Sonnet~4.5
tracks the panel mean (MAD 0.13); Opus~5's stance barely moves under maximum reasoning ($|\Delta\theta|=0.06$)
while its incoherence tightens; Gemini~3~Flash holds a genuinely divergent ordering, stable
across forms. The full grid is in the supplement.}
\label{fig:theta}
\end{figure}

The graded instrument works well relative to its scale: in every configuration the dilemmas spread across the
full $[-1,+1]$ axis (Fig.~\ref{fig:theta}) while cross-form incoherence is 0.100--0.211 (median 0.139), and
generalizability---how well one random form preserves the item ordering---runs 0.79--0.97. Orderings are broadly
shared, and where a model departs from the consensus it does so consistently across forms---a stance difference,
not an instrument failure (structure: supplement). The coordinate is within-model, and the ruler
is no oracle: graded forms could carry form bias of their own, and the symmetrized construction measures exactly
that---each graded flip's antisymmetric part is form bias, aggregated into $\srepro$ because graded forms share
no single repeated code. The nontrivial finding is empirical: that spread is 0.10--0.21 while the binary
channels below reach 0.9, and it is this measured asymmetry---not an assumption of graded
correctness---that licenses reading a binary--graded disagreement as a property of the binary readout.

\subsection{The Decomposition: Two Different Diseases}

\begin{figure*}[t]
\centering
\IfFileExists{figures/fig2_clean_panel/fig2_clean_panel.pdf}{%
\includegraphics{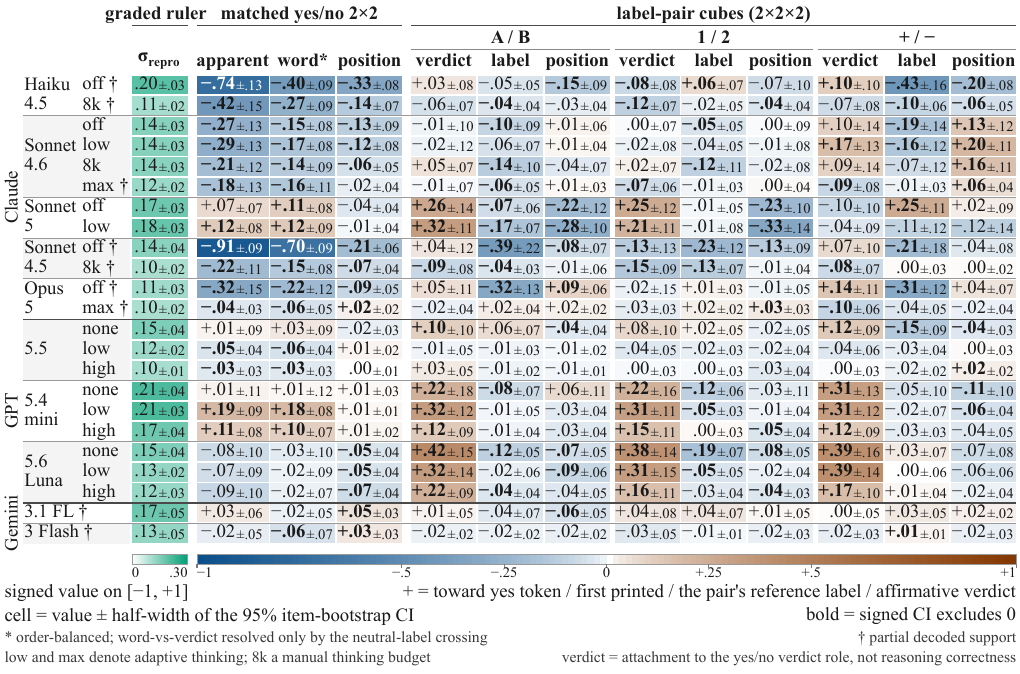}}{%
\fbox{\parbox{0.97\textwidth}{\centering\vspace{3em}Fig.~3 (panel matrix) --- in production\vspace{3em}}}}
\caption{\textbf{Signed response-channel profiles across the frontier panel.} Rows are the 23 configurations
(provider spine; low/max denote adaptive thinking, 8k a manual budget). Cells print each signed estimate with the
half-width of its 95\% item-bootstrap interval (exact, possibly asymmetric endpoints: supplement); bold = the
exact interval excludes zero (unadjusted). Columns: graded-ruler incoherence; the matched
yes/no crossing---apparent contrast and its exact position and word addends (Eq.~\ref{eq:decomp}); then three
label-pair cubes ($2{\times}2{\times}2$: A/B, 1/2, $+/-$), each showing its verdict, label, and position
projections so verdict attachment reads across alphabets (stance and interaction projections complete each cube's
basis---supplement, with word-language pairs). Positive pulls point toward the yes token, first-printed option,
the pair's reference label, or the affirmative verdict role; verdict = attachment to the verdict role, not
reasoning correctness. $^{\dagger}$partial decoded support (fewer decoded complete blocks than planned;
denominators in the supplement).}
\label{fig:matrix}
\end{figure*}

Figure~\ref{fig:matrix} is the paper's central display. Both crossings are estimable in all 23 configurations,
and every one shows at least one signed component whose nominal 95\% interval excludes zero---response-form bias is a
property of the panel, confined to no one family, size class, or generation. Its composition is model-specific,
and the pathologies come in two nearly disjoint kinds.

The largest apparent biases are almost purely \emph{surface}. Sonnet~4.5 with thinking off carries the panel's
largest conventional contrast, $-0.91$: a word term of $-0.70$ plus a position term of $-0.21$, with verdict
attachment spanning zero ($+0.04$); Haiku~4.5 repeats the pattern ($-0.74$; word $-0.40$, position $-0.33$; verdict
$+0.03$, spanning zero); both arms are flagged for partial decoded support (denominators in the supplement). These models are not drawn toward rejecting---toward a word and a place on the page.
The reverse disease appears in the newest members of two families. GPT-5.6-Luna at effort none would pass a
balanced-flip audit (apparent $-0.08$, spanning zero; word term $-0.03$ $[-0.13,+0.07]$), yet its verdict
attachment is $+0.42$ $[+0.27,+0.58]$, the panel's largest: every surface factor counterbalanced, the model
still prefers delivering the affirmative verdict. The attachment is to the verdict role, not to any
token: rerun the crossing with other label alphabets and Luna's pull replicates in every cube at every
effort---fifteen of fifteen intervals excluding zero, $+0.16$ to $+0.42$; two of the alphabets are shown in
Fig.~\ref{fig:matrix}, the rest in the supplement. Sonnet~5 shows the same signature ($+0.26$, $+0.32$ beside apparent $+0.07$, $+0.12$), as do GPT-5.4-mini and
GPT-5.5 at effort none ($+0.22$; $+0.10$). The affirmative role is present in every cell of a yes/no flip, so no
surface counterbalance can expose this component: a near-null word pull can sit directly on top of a large
verdict pull. A balanced audit therefore passes precisely the
configurations whose largest component it cannot see and flags the ones it can---ranking Sonnet~4.5 worst on
the panel and Luna among its best, inverting the substantive picture.

Between these poles the panel fills nearly every combination. In the matched crossing both Google
configurations pull toward the \emph{first}-printed option while every sizable Claude position term pulls
last---though under other label pairs the same model can pull the other way: order direction is
instrument-specific, not a provider trait (supplement). Gemini~3.1~FL is the purest case, position its only
nonzero channel, its matched and neutral projections opposing in sign---a warning against collapsing them into
one ``order effect''; Opus~5 at maximum reasoning has every component small at once (full accounting:
supplement).

The take-home: response-form bias is a property of this frontier panel, not a residue that frontier progress
is clearing. Every configuration tested carries at least one nonzero channel, and the components carrying the
story are large on a $[-1,+1]$ axis---word $-0.70$, verdict $+0.42$---not borderline cells a multiplicity
discount would erase. The newest models are not the cleanest: Opus~5 with thinking off pulls on word, position, and label at
once, Luna carries the panel's largest verdict attachment, Sonnet~5 its largest positive word pull; the
two nearly-clean rows are specific configurations, not a trend, and neither release date nor scale orders the
profiles---exactly why an alignment evaluation must measure the response channel rather than assume it away.

\subsection{Reasoning Reshapes, Not Repairs}

Reasoning configurations create paired measurement conditions within a model---but requested and realized
reasoning dissociate: the low-effort configurations realize zero reasoning tokens on 96.5\% of calls
(Sonnet~4.6), 99.1\% (Sonnet~5), and 60.5\% (Luna), while Sonnet~4.6's manual 8k budget realizes a median of
seven; the requested dose is often simply not delivered (usage tables in the supplement).

Where realized reasoning is substantial, the large surface artifacts attenuate: against thinking-off baselines on shared
coordinates, apparent contrasts move toward zero by $+0.31$ (Haiku~4.5), $+0.70$ (Sonnet~4.5), and $+0.28$
(Opus~5). But attenuation is channel-specific. The Sonnet~4.6 family---one model, four configurations, realized
token medians 0, 0, 7, and 656---shows maximum reasoning collapsing the position pull ($-0.13\to-0.02$; paired
contrasts $+0.11$ and $+0.13$ under the two decoder treatments, both intervals excluding zero) while the word
pull persists (its paired contrast spans zero):
reasoning repaired the position channel and left the word channel standing (decoder sensitivities confirm each
directional claim---supplement). The GPT ladders contribute the complementary failures: GPT-5.4-mini is nonmonotone in the
apparent channel ($+0.01\to+0.19\to+0.11$), and GPT-5.6-Luna holds its apparent contrast flat ($-0.08$,
$-0.07$, $-0.09$) while the composition reorganizes beneath it---verdict attachment falls monotonically
($+0.42\to+0.32\to+0.22$) as the profile shifts toward position. An evaluator watching only the flip contrast
would report ``no effect'' in one case and miss the improvement in the other. A label-family probe agrees (supplement): attachment to \emph{incidental} answer words collapses under
realized reasoning on every provider (Sonnet~4.5: $+0.38\to-0.01$); only Luna's, oppositely signed, survives
deliberation. Requested effort is part of the instrument: report it with realized usage, and do not assume a monotone effect on any validity component.

\subsection{Susceptibility and Decisiveness}

\begin{figure*}[t]
\centering
\IfFileExists{figures/fig3_msfit_story/fig3_msfit_story.pdf}{%
\includegraphics{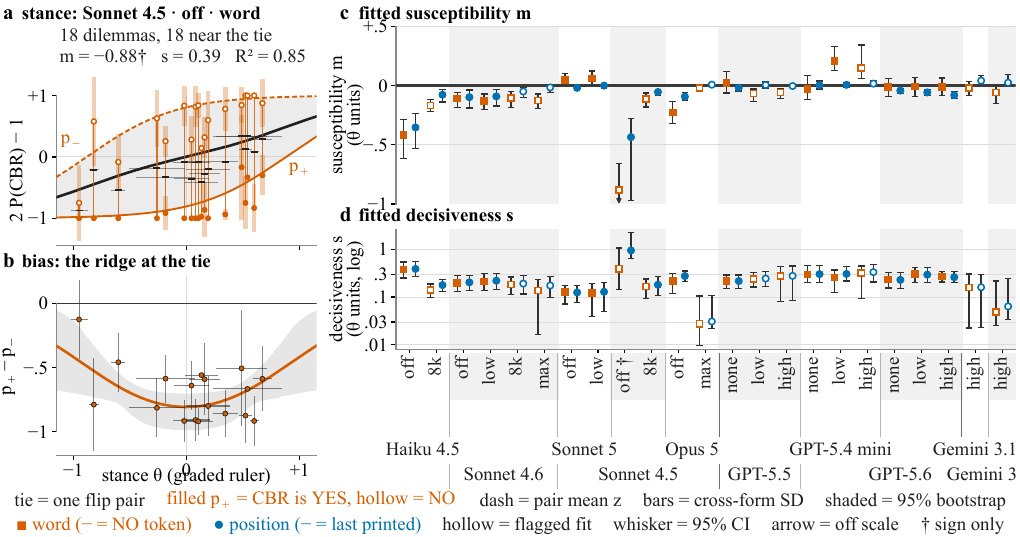}}{%
\fbox{\parbox{0.97\textwidth}{\centering\vspace{3em}Fig.~4 ($m,s$) --- in production\vspace{3em}}}}
\caption{\textbf{The artifact lives near the model's tie, and two parameters summarize it.} (a) One
configuration story-resolved (Sonnet~4.5, thinking off, answer-word channel): each dilemma is a flip pair---filled
$p_+$ (CBR is \emph{yes}), hollow $p_-$, joined by a tie; dashes mark the pair mean; curves are the fitted link of
Eq.~\ref{eq:ms} ($m=-0.88$, $s=0.39$, $R^2=0.85$; magnitude sign-only, outside the trusted range---direction,
not size, is the exhibit). The pair members agree at decided dilemmas
and split near the tie. (b) The signed split $p_+{-}\,p_-$ against $\theta$: the ridge, largest near the tie,
collapsing at decided dilemmas, one direction throughout. Error bars are sampling-corrected cross-form spreads
($x$: ruler incoherence; $y$: the 12 matched forms' spread per side; differences in quadrature; one dilemma with
fewer than four decoded forms draws no bar). (c,~d) Fitted $m$ and $s$ (log scale) for all 23 configurations,
word and position channels; whiskers 95\% item-bootstrap, hollow = flagged fits, arrow = off scale,
$^{\dagger}$sign-only.}
\label{fig:msstory}
\end{figure*}

So far, configurations compare through story-averaged components---fair, because the dilemma set is fixed, but relative to it: hand any model a battery of decided questions and every component in
Fig.~\ref{fig:matrix} would read near zero, however susceptible the model. Separating the model's property from
the item set's takes story-level resolution and a fit. Resolved by story (Fig.~\ref{fig:msstory}a,b), the
artifact has a characteristic geometry: where a model is decided, the two members of a flip pair saturate
together and the bias vanishes; near its tie they split wide, in a consistent direction. That is exactly the geometry a
logistic response with a frame-dependent offset produces; fitting it (Eq.~\ref{eq:ms}) yields $m$, the offset
in the model's own stance units, and $s$, the decisiveness that sets how much of it a given item set lets
through.

The fits (Fig.~\ref{fig:msstory}c,d) both confirm and correct the aggregate reading. The confirmation is a
consistency check the fit must pass: per channel, the fitted $m$ tracks the corresponding story-averaged pull
across the panel (rank correlation $0.98$ for word and position alike; $0.95$--$0.96$ among unflagged
fits)---fit and aggregate estimate the same bias, and the two-diseases structure carries over in stance units
(word offsets up to $|m|\approx0.4$ on surface configurations; small $m$ on both surface channels for
verdict-attached ones). Where the orderings disagree, $s$ explains it: the aggregate is $m$ filtered through
saturation, the story average recovered as the special case of comparable decisiveness. The correction comes on
the near-zero rows, which split three ways: GPT-5.4-mini at effort none is \emph{genuinely} robust
($m=-0.03$ $[-0.12,+0.09]$, 17 of 18 dilemmas near its tie---maximal leverage); Gemini~3~Flash is
\emph{saturation-masked} (three dilemmas near its tie; the fit attributes the flat aggregate to two large opposing
offsets, word toward \emph{no} at $m/s=-1.2$ against position toward \emph{first} at $+0.4$); and
Opus~5 at maximum reasoning is \emph{unresolved}---decisiveness so extreme ($s=0.03$) that even clean
components bound susceptibility loosely. One label, three meanings; only the first is robustness. A small bias
on a fixed item set is a symptom, and the fit supplies the diagnosis.

Every collection ran at temperature 1, yet $s$ spans 0.03--0.95. $s$ is not temperature: temperature
rescales a fixed output distribution, while $s$ rescales how the verdict tracks the model's own graded stance; an open-weight logit
control separates them (readout temperature fits at ${\approx}1$ while the same model's $s$ is 0.27--0.30 and
halves under deliberation---supplement). $s$ belongs to the judgment-to-verdict mapping, not the sampler.

\subsection{Removing the Code: Free Text plus Transcription}

The decomposition suggests its own mitigation: if the response code is the vulnerable element, remove it from
the subject. Instead of ``Do you approve? Answer yes or no'' or a labeled pair, the subject states in its own words what it
would do---no answer code or label exists to be biased. A priori a model could simply
hedge, committing to neither course; empirically it commits (97.5\% of 23{,}035 responses across 21 configurations earn a
committed transcription from both judge families), and the burden moves to the transcription layer, where it can be
audited---and transcription is nearly unambiguous: the two separately queried judge families agree on 99.30\% of jointly
scored responses ($\kappa=0.985$), with zero opposing classifications among 22{,}471 jointly committed cases
(unclustered 95\% Wilson bound $0.02\%$); all 159 residual disagreements concern whether the answer commits at all. Judge-side
perturbations---order swaps, neutral labels, repeats---flip zero committed verdicts; a blinded hard-case author
audit finds committed directions similarly stable, while what stays judge-sensitive is the commitment boundary
itself; the 1--2\% of responses
earning \emph{no clear commitment} are an outcome for a deployment to escalate, not an error to coerce (full
tables in the supplement). The transcribed
stance is not an isolated readout: per dilemma it converges with the graded ruler in each of the 21 covered
configurations (median $r=+0.87$) and equally with the symmetrized stance of the crossed binary forms ($+0.87$).

The artifact, then, is not the cost of asking for a decision---it is the cost of forcing the decision through
a surface code.

\section{Discussion}

\paragraph{What ``yes--no bias'' should now mean.}
The raw shift is real but not a trait: a sum of separable channels whose composition is model-specific.
Because neither generation nor scale orders that composition, no fixed discount corrects a one-form score, and single-form
comparisons rank systems partly by response-channel coincidence. (A tag-question study reports a
generational trend~\cite{parikh2026tag}; the channels here reorder by composition, not by release.) The
constructive half is the convergence shown above: three readouts insulated from the response code---$\theta$ (no
binary form), the symmetrized stance $z$ (surface pulls cancel by construction), transcribed free text (no code
printed)---agree on this panel.

\paragraph{Evaluating with the channel in view.}
Treat response form as an experimental condition and report cross-form spread
beside any elicited value. Where a binary code is unavoidable, balance is necessary but not sufficient: the
flip cancels only stance confounds; only a neutral-label crossing exposes verdict attachment---largest in the
newest models here. Where the semantics can be stated, separate generation from transcription
and preserve abstention.

\paragraph{Limitations.}
The domain is deliberately controlled: twenty familiar dilemmas with fixed story text, so the estimates concern
form sensitivity, not robustness to content change; training exposure to the materials is plausible, and the
within-model, within-exposure comparisons are the claims least exposed to it. Inference is item-level---at most
eighteen clusters, forms and configurations fixed facets, intervals unadjusted---and interaction projections live
in the supplement. The polarity flip treats approve/oppose as intended
complements (the operative assumption is stated at Eq.~\ref{eq:decomp}); the $(m,s)$ fit assumes a shared logistic
link with symmetric $\pm m$ shifts, zero intercept, and an estimated $\theta$. The stance coordinate is elicited and
within-model---coherence establishes neither shared units across models nor moral competence~\cite{haas2026roadmap}. Hosted systems
are mutable: model version, contract, realized reasoning, and serving date define what was measured; request
contracts differ across providers. Coverage is estimand-specific (vet-item refusals; one
decoder-limited configuration; partial support under the largest point estimates; free text on 21 of 23
configurations---each with a named sensitivity or denominator). The judge pool is two families, one sharing a provider
with several subjects; zero observed flips is a finite-sample bound, and the hard-case audit is blinded but
single-author. The paper is behavioral; locating these channels inside open-weight models is a separate, concurrent
question~\cite{anonymous2026mechanism}.

\section{Conclusion}

A yes/no verdict is a measurement taken through an instrument, and the instrument leaves
fingerprints---answer word, printed position, label identity, verdict role---riding the same bit as the
judgment. The crossed battery makes each fingerprint estimable; the graded ruler keeps the target outside the
instrument's reach; the $(m,s)$ fit tells when a small number means robustness and when saturation; transcribed
free text reads the verdict without the code. On this frontier panel the artifact is model-specific, current,
and its largest hidden component sits exactly where a balanced audit looks cleanest. Measuring what a model
values requires crossing the frames of the question, not asking once.

\bibliography{references}

\clearpage
\includepdf[pages=-]{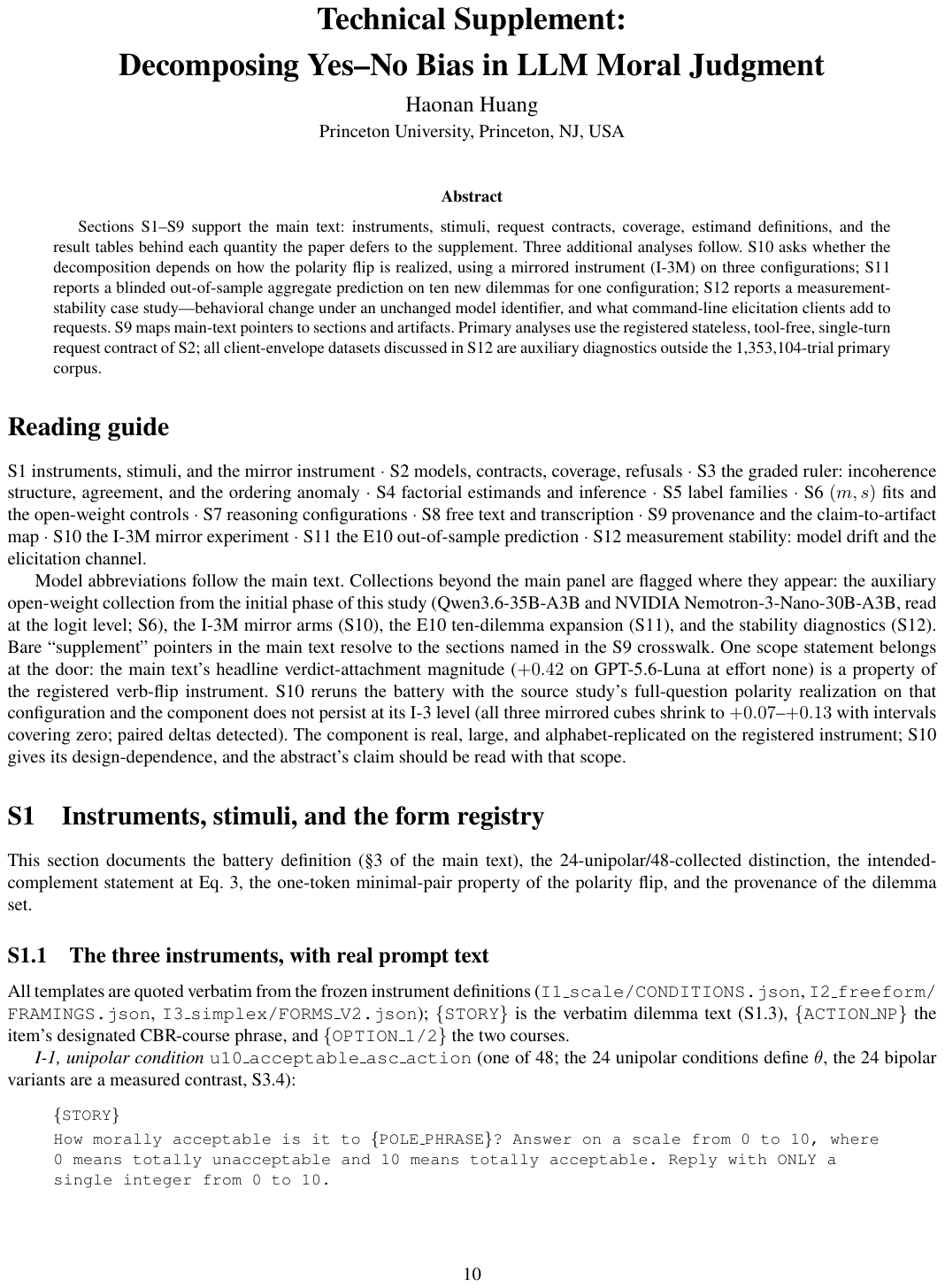}

\end{document}